%% file: main.tex
\documentclass[runningheads]{llncs}

\usepackage[mobile]{eccv}
\usepackage{eccvabbrv}
\usepackage{graphicx}
\usepackage{booktabs}
\usepackage{threeparttable}
\usepackage{xspace}
\usepackage{siunitx}
\usepackage{graphicx}
\usepackage{amssymb}
\usepackage{threeparttable}
\usepackage{color}
\usepackage[normalem]{ulem}
\usepackage{multirow}
\usepackage{float}
\usepackage{amsfonts}
\usepackage{bm}
\usepackage{array}
\usepackage{overpic}
\usepackage{enumitem}
\usepackage{colortbl}
\usepackage{algorithm}
\usepackage{algorithmicx}
\usepackage{algpseudocode}
\usepackage{cite}
\usepackage[export]{adjustbox}
\usepackage{pifont}
\usepackage[accsupp]{axessibility}
\usepackage{orcidlink}
\usepackage{hyperref}
\usepackage{amsfonts}
\usepackage{bbm}
\usepackage{comment}
\usepackage{tcolorbox}
\usepackage{tabulary}
\usepackage[frozencache=true,cachedir=minted-cache]{minted}
\newcolumntype{I}{!{\vrule width 1pt}}
\newcolumntype{x}[1]{>{\centering\arraybackslash}p{#1pt}}
\newcolumntype{y}[1]{>{\raggedright\arraybackslash}p{#1pt}}
\newcolumntype{z}[1]{>{\raggedleft\arraybackslash}p{#1pt}}

\makeatletter
\newcommand{\thickhline}{
    \noalign {\ifnum 0=`}\fi \hrule height 1pt
    \futurelet \reserved@a \@xhline
}
\makeatother

\newcommand{\pub}[1]{\color{gray}{\tiny{[{#1}]}}}

\usepackage{caption}
\begin{document}

\title{\emph{VideoAgent}: Long-form Video Understanding with Large Language Model as Agent}
\titlerunning{\emph{VideoAgent}}

\author{Xiaohan Wang$^\star$ \and Yuhui Zhang$^\star$ \and Orr Zohar \and Serena Yeung-Levy}
\authorrunning{Wang et al.}
\institute{Stanford University \\
\email{\{xhanwang,yuhuiz,orrzohar,syyeung\}@stanford.edu}}

\renewcommand{\thefootnote}{\ensuremath{\dagger}}

{
  \renewcommand{\thefootnote}
    {\fnsymbol{footnote}}
  \footnotetext[1]{Equal contribution. Project page: \scriptsize{\url{https://wxh1996.github.io/VideoAgent-Website/}}.}
}

\maketitle

\input{secs/0_abs}
\input{secs/1_intro}
\input{secs/2_related}
\input{secs/3_method}
\input{secs/4_results}
\input{secs/5_conclusion}

\bibliographystyle{splncs04}
\bibliography{main}

\appendix
\input{secs/X_suppl}

\end{document}

%% file: secs/0_abs.tex
\begin{abstract}
Long-form video understanding represents a significant challenge within computer vision, demanding a model capable of reasoning over long multi-modal sequences. Motivated by the human cognitive process for long-form video understanding, we emphasize interactive reasoning and planning over the ability to process lengthy visual inputs. We introduce a novel agent-based system, \emph{VideoAgent}, that employs a large language model as a central agent to iteratively identify and compile crucial information to answer a question, with vision-language foundation models serving as tools to translate and retrieve visual information. Evaluated on the challenging EgoSchema and NExT-QA benchmarks, \emph{VideoAgent} achieves 54.1\% and 71.3\% zero-shot accuracy with only 8.4 and 8.2 frames used on average. These results demonstrate superior effectiveness and efficiency of our method over the current state-of-the-art methods, highlighting the potential of agent-based approaches in advancing long-form video understanding.
\keywords{Long-form Video Understanding \and Large Language Model Agent \and Vision-Language Foundation Models }
\end{abstract}

%% file: secs/1_intro.tex
\section{Introduction}
\label{sec:intro}

\begin{figure}[tb]
  \centering
  \includegraphics[width=\linewidth]{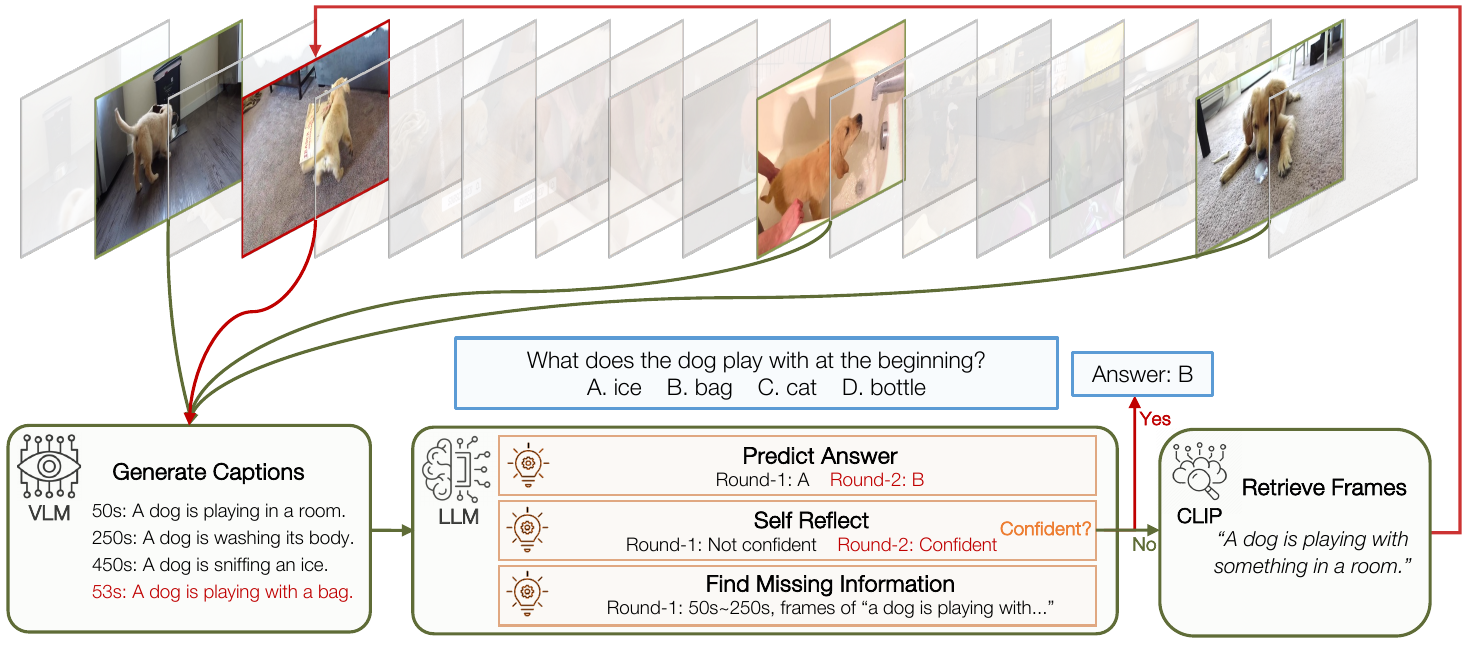}
  \vspace{-1.5em}
  \caption{\emph{Overview of VideoAgent.} Given a long-form video, \emph{VideoAgent} iteratively searches and aggregates key information to answer the question. The process is controlled by a large language model (LLM) as the agent, with the visual language model (VLM) and contrastive language-image model (CLIP) serving as tools.}
  \vspace{-1em}
  \label{fig:overview}
\end{figure}

Understanding long-form videos, which range from minutes to hours, poses a significant challenge in the field of computer vision. This task demands a model capable of processing multi-modal information, handling exceedingly long sequences, and reasoning over these sequences effectively.

Despite numerous attempts~\cite{sun2022long, ma2023vista, wu2022memvit, song2023moviechat,jin2023chatunivi,hussein2019videograph, wang2021supervoxel, yang2020gives, nguyen2022s4nd,islam2022long} to address this challenge by enhancing these capabilities, existing models struggle to excel in all three areas simultaneously. Current large language models (LLMs) excel in reasoning and handling long contexts~\cite{wei2022chain,yao2024tree,touvron2023llama,jiang2024mixtral}, yet they lack the capability to process visual information. Conversely, visual language models (VLMs) struggle to model lengthy visual inputs~\cite{videoLlava,liu2023llava,li2023blip,li2023blip2,hong2023cogagent}. Early efforts have been made to enable VLMs' long context modeling capability, but these adaptations underperform in video understanding benchmarks and are inefficient in dealing with long-form video content~\cite{liu2024world}.

Do we really need to feed the entire long-form video directly into the model? 
This diverges significantly from how humans achieve the long-form video understanding task.
When tasked with understanding a long video, humans typically rely on the following interactive process to formulate an answer:
The process begins with a quick overview of the video to understand its context.
Subsequently, guided by the specific question at hand, humans iteratively select new frames to gather relevant information. 
Upon acquiring sufficient information to answer the question, the iterative process is concluded, and the answer is provided. 
Throughout this process, the reasoning capability to control this iterative process is more critical than the capacity to directly process lengthy visual inputs.

Drawing inspiration from how humans understand long-form videos, we present VideoAgent, a system that simulates this process through an agent-based system. We formulate the video understanding process as a sequence of states, actions, and observations, with an LLM serving as the agent controlling this process (Figure~\ref{fig:overview}). Initially, the LLM familiarizes itself with the video context by glancing at a set of uniformly sampled frames from the video. During each iteration, the LLM assesses whether the current information (state) is sufficient to answer the question; if not, it identifies what additional information is required (action). Subsequently, it utilizes CLIP~\cite{radford2021learning} to retrieve new frames containing this information (observation) and VLM to caption these new frames into textual descriptions, updating the current state. This design emphasizes the reasoning capability and iterative processes over the direct processing of long visual inputs, where the VLM and CLIP serve as instrumental tools to enable the LLM to have visual understanding and long-context retrieval capabilities.

Our work differs from previous works in two aspects. Compared to the works that uniformly sample frames or select frames in a single iteration~\cite{yu2023self, xu2023retrievalbased, lei2021less}, our method selects frames in a multi-round fashion, which ensures the information gathered to be more accurate based on the current need. Compared to the works that retrieve frames using the original question as the query~\cite{yu2023self, xu2023retrievalbased}, we rewrite the query to enable more accurate and fine-grained frame retrieval.

Our rigorous evaluation of two well-established long-form video understanding benchmarks, EgoSchema~\cite{mangalam2023egoschema} and NExT-QA~\cite{xiao2021next}, demonstrates \emph{VideoAgent}'s exceptional effectiveness and efficiency compared to existing methods. \emph{VideoAgent} achieves 54.1\% and 71.3\% accuracy on these two benchmarks, respectively, outperforming concurrent state-of-the-art method LLoVi~\cite{zhang2023simple} by 3.8\% and 3.6\%. Notably, \emph{VideoAgent} only utilizes 8.4 frames on average to achieve such performance, which is 20x fewer compared to LLoVi. Our ablation studies highlight the significance of the iterative frame selection process, which adaptively searches and aggregates relevant information based on the complexity of the videos. Additionally, our case studies demonstrate that \emph{VideoAgent} generalizes to arbitrarily long videos, including those extending to an hour or more.

In summary, \emph{VideoAgent} represents a significant stride for long-form video understanding, which embraces the agent-based system to mimic human cognitive process and underscores the primacy of reasoning over long-context visual information modeling. We hope our work not only sets a new benchmark in long-form video understanding but also sheds light on future research in this direction.

%% file: secs/2_related.tex
\section{Related Work}

\subsection{Long-form Video Understanding}

Long-form video understanding is a particularly challenging domain in computer vision due to the inherent complexity and high dimensionality of spatio-temporal inputs, which leads to significant computational demands. Long-form video understanding methods need to balance computational efficiency and performance, and can broadly be categorized into selective or compressive sparsity strategies.  

Compressive sparsity methods~\cite{yeung2016end, sun2022long, wu2022memvit, song2023moviechat,jin2023chatunivi,hussein2019videograph, wang2021supervoxel, yang2020gives, nguyen2022s4nd,islam2022long}, attempt to compress the video into meaningful embeddings/representations with the minimum possible dimensionality. For example, MovieChat~\cite{song2023moviechat} employs a memory consolidation mechanism that merges similar adjacent frame tokens based on cosine similarity, effectively reducing token redundancy in long video sequences. Chat-UniVi~\cite{jin2023chatunivi} utilized kNN clustering to spatio-temporally compress video tokens. However, the compression need not happen on the embeddings themselves, and can be compressed into space-time graphs~\cite{hussein2019videograph, wang2021supervoxel, yang2020gives} or even text~\cite{zhang2023simple, romero2024questioninstructed,li2023videochat}. For example, Zhang et. al.~\cite{zhang2023simple} introduced LLoVi, and have shown that simply captioning videos before and prompting an LLM with these captions can serve as a strong baseline.

Meanwhile, selective-compressive methodologies attempt to sub-sample the video into more meaningful frames, utilizing the input question/text as a guide, and in effect attempt to only sample frames relevant to the question at hand~\cite{yu2023self,gao2023mist,ren2023timechat,li2024llms, xu2023retrievalbased}. For instance, methods such as R-VLM and R2A~\cite{xu2023retrievalbased, pan2023retrievingtoanswer, han2023sas} utilize a CLIP model to retrieve relevant frames given a text prompt, and while Q-ViD~\cite{romero2024questioninstructed} utilize the question to selectively caption the video.
Unlike previous works, we allow the LLM to direct the video frames to be sampled by the captioner.

\subsection{LLM Agents}

An agent is defined as an entity that makes decisions and takes actions in a dynamic, real-time environment to achieve some specific goals. Advances in large language models (LLMs), especially their emerging reasoning and planning capabilities~\cite{wei2022chain,zhou2022least,yao2024tree}, has inspired recent research in natural language processing to leverage them as agents in real-world scenarios~\cite{yao2022react,park2023generative}. These models have demonstrated great success across various scenarios, such as online search, card game playing, and database management~\cite{liu2023agentbench,ma2024agentboard,yao2022webshop}. Their effectiveness is further amplified with methods such as chain-of-thought reasoning or self-reflection~\cite{wei2022chain,shinn2024reflexion}.

Simultaneously, the computer vision community has begun to explore LLM-as-agent-based approach in diverse visual contexts, such as GUI understanding and robot navigation~\cite{suris2023vipergpt,hong2023cogagent,driess2023palm,brohan2023rt}. In the area of long-form video understanding, several studies have made an initial attempt with an agent-like approach, which utilize LLMs to interact with external tools or to incorporate additional functionalities~\cite{suris2023vipergpt,gao2023assistgpt,yang2024doraemongpt}. Unlike these approaches, our work reformulates video understanding as a decision-making process, which is inspired by how humans solve video interpretation methods. We view the video as an environment where decisions involve either seeking more information or concluding the interaction. This perspective has guided the creation of \emph{VideoAgent}, a novel framework that significantly diverges from existing methodologies by emphasizing the decision-making aspects inherent in video understanding.

%% file: secs/3_method.tex
\section{Method}

\begin{figure}[!tb]
    \centering
    \includegraphics[width=\linewidth]{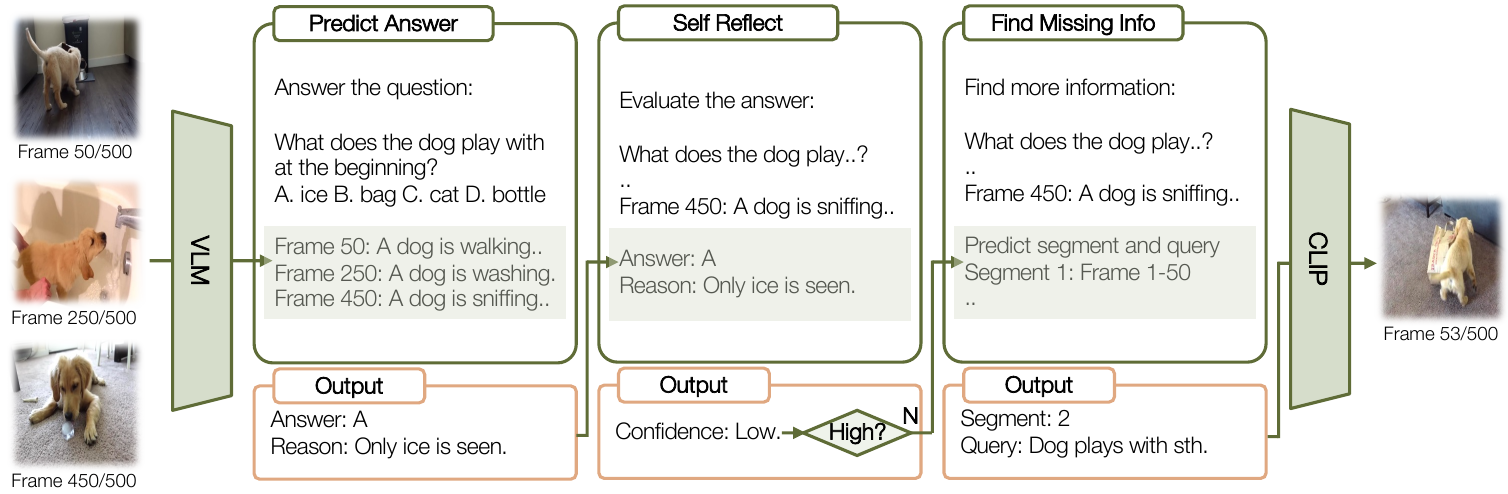}
    \vspace{-1.7em}
    \caption{\emph{Detailed view of VideoAgent's iterative process.} Each round starts with the state, which includes previously viewed video frames. The large language model then determines subsequent actions by answering prediction and self-reflection. If additional information is needed, new observations are acquired in the form of video frames.}
    \vspace{-1.5em}
    \label{fig:method}
\end{figure}

In this section, we introduce the method of \emph{VideoAgent}. \emph{VideoAgent} is inspired by the human cognitive process of understanding long-form videos. Given a video with the question, a human will first glance at several frames to understand its context, then iteratively search additional frames to obtain enough information to answer the question, and finally aggregate all the information and make the prediction. 

We formulate the process into a sequence of states, actions, and observations $\{(s_t, a_t, o_t)| 1\le t \le T\}$, where the state is the existing information of all the seen frames, action is whether to answer the question or continue to search new frames, observation is the new frames seen in the current iteration, and $T$ is the maximum number of iterations. 

We leverage large language model (LLM) GPT-4~\cite{openai2023gpt4} as an agent to perform the above process (Figure~\ref{fig:overview}). LLMs have been demonstrated to have memory, reasoning and planning, and tool-use capability~\cite{tay2020long,wei2022chain,zhou2022least,schick2024toolformer}, which can be used to model states, actions, and observations, respectively.

\subsection{Obtaining the Initial State}

To start the iterative process, we first familiarize the LLM with the context of the video, which can be achieved by glancing at $N$ frames uniformly sampled from the video. Since the LLM has no capability for visual understanding, we leverage vision-language models (VLMs) to convert the visual content to language descriptions. Specifically, we caption these $N$ frames with the prompt ``describe the image in detail'' and feed the captions to the LLM. This initial state $s_1$ records a sketch of the content and semantics of the video.

\subsection{Determining the Next Action}

Given the current state $s_t$ that stores the information of all the seen frames, there are two possible options for the next action $a_t$:

\begin{itemize}
    \item \textbf{Action 1: answer the question.} If the information in state $s_t$ is enough to answer the question, we should answer the questions and exit the iterative process.
    \item \textbf{Action 2: search new information.} If the current information in $s_t$ is insufficient, we should decide what further information is required to answer the question and continue searching for it.
\end{itemize}

To decide between actions 1 and 2, we need the LLM to reason over the question and existing information. This is achieved by a three-step process. First, we force the LLM to make a prediction based on the current state and question via chain-of-thought prompting. Second, we ask the LLM to self-reflect and generate a confidence score based on the state, question, prediction and its reasoning process generated by step 1. The confidence score has three levels: 1 (insufficient information), 2 (partial information), and 3 (sufficient information). Finally, we choose action 1 or 2 based on the confidence score. This process is illustrated in Figure~\ref{fig:method}.
We propose to use a three-step process over a single-step process that directly chooses action as direct prediction always decides to search for new information (Action 2). This self-reflection process is motivated by~\cite{shinn2024reflexion}, which has demonstrated superior effectiveness in natural language processing.

\subsection{Gathering a New Observation}

Suppose the LLM determines insufficient information to answer the question and chooses to search for new information. In that case, we further ask the LLM to decide what extra information is needed so that we can leverage tools to retrieve them (Figure~\ref{fig:method}). Since some piece of information could occur multiple times within a video, we perform segment-level retrieval instead of video-level retrieval to enhance the temporal reasoning capability. For example, suppose the question is ``What is the toy left on the sofa after the boy leaves the room?'' and that we have seen the boy leave the room at frame $i$. If we retrieve with the query ``a frame showing the toy on the sofa,'' there may be frames before frame $i$ containing ``toy on the sofa'', but they are irrelevant to answering the question.

To perform segment-level retrieval, we first split the video into different segments based on the seen frame indices, and ask the LLM to predict what segments to retrieve with the query texts. For example, if we have seen frames $i$, $j$, and $k$ of a video, one valid prediction is segment 2 (frame $i$ to $j$) with the query ``a frame showing the toy on the sofa''.

We leverage CLIP~\cite{radford2021learning} to obtain this additional information given the output by the LLM. Specifically, given each query and segment, we return the image frame with the highest cosine similarity with the text query in that segment. These retrieved frames are served as observations to update the state. 

The use of CLIP in the retrieval step is computationally efficient and negligible compared to using an LLM or VLM for several reasons. Firstly, CLIP's feature computation involves just a single feed-forward process. Secondly, CLIP employs an image-text late interaction architecture, enabling the caching and reusing of image frame features across different text queries. Lastly, our segment-level retrieval design only requires computing features within specific segments, further enhancing efficiency. Empirically, our experiments show that CLIP computations are less than 1\% of that of a VLM and LLM.

\subsection{Updating the Current State}

Finally, given the new observations (i.e., retrieved frames), we leverage VLMs to generate captions for each frame, and then simply sort and concatenate the new captions with old frame captions based on frame index, and ask the LLM to generate next-round predictions. 

A question one may posit is why we leverage the multi-round process, as some existing works use all or uniformly sampled frames as the state in a single step~\cite{lei2021less,zhang2023simple}. There are many advantages of our approach over these baselines. First, using too many frames introduces extensive information and noise, which leads to performance degradation because LLMs suffer from long contexts and can be easily distracted~\cite{shi2023large,liu2024lost}. Furthermore, it is computationally inefficient and hard to scale up to hour-long videos due to the LLM context length limit~\cite{openai2023gpt4}. On the opposite, using too few frames may not capture relevant information. Our adaptive selection strategy finds the most relevant information and requires the lowest cost to answer questions at different difficulty levels.

We summarize the \emph{VideoAgent} as Algorithm~\ref{alg}.
\vspace{-1em}

\begin{algorithm}[htbp]
\caption{\emph{VideoAgent}}
\begin{algorithmic}[1]
\Require Video $v$, question $q$, LLM $F_l$, VLM $F_v$, CLIP $F_c$, max iteration $T$, confidence threshold $C$
\Ensure Prediction $\hat{y}$, state-action-observation sequence $\{s_t, a_t, o_t | 1 \le t \le T\}$

\State $s_1 \gets \texttt{GenerateCaptions}(F_v, \texttt{UniformSample}(v))$
\For{$t = 1$ \textbf{to} $T$}
\State $\hat{y} \gets \texttt{PredictAnswer}(F_l, s_t, q)$
\State $c \gets \texttt{SelfReflect}(F_l, s_t, q, \hat{y})$
\If{$a_t \gets \mathbbm{1}_{[c \ge C]}$}
\State {\textbf{break}}
\Else

\State $h \gets \texttt{FindMissingInfo}(F_l, s_t, q)$
\State $o_t \gets \texttt{RetrieveFrames}(F_c, v, h)$
\State $s_{t+1} \gets \texttt{Merge}(s_t, \texttt{GenerateCaptions}(F_v, o_t))$
\EndIf
\EndFor
\State \Return $\hat{y}$, $\{s_t, a_t, o_t | 1 \le t \le T\}$
\end{algorithmic}
\label{alg}
\end{algorithm}
\vspace{-1em}

%% file: secs/4_results.tex
\section{Experiments}

In this section, we first introduce the datasets and implementation details, and then we present the results, analyses, ablations, and case studies of \emph{VideoAgent}.

\subsection{Datasets and Metrics}

In our experiments, we use two distinct well-established datasets to benchmark our model's performance, with a particular focus on zero-shot understanding capabilities.

\paragraph{EgoSchema~\cite{mangalam2023egoschema}.} EgoSchema is a benchmark for long-form video understanding, featuring 5,000 multiple-choice questions derived from 5,000 egocentric videos. These videos provide an egocentric viewpoint of humans engaged in a wide range of activities. A distinctive feature of this dataset is the length of its videos, each lasting 3 minutes. EgoSchema comprises only a test set, with a subset of 500 questions having publicly available labels. The full set of questions is evaluated exclusively on the official leaderboard.

\paragraph{NExT-QA~\cite{xiao2021next}.} The NExT-QA dataset includes 5,440 natural videos that feature object interactions in daily life, accompanied by 48,000 multiple-choice questions. The average length of video is 44 seconds. These questions fall into three categories: Temporal, Causal, and Descriptive, providing a comprehensive evaluation for video understanding models. In line with standard practices, our zero-shot evaluation focused on the validation set, which contains 570 videos and 5,000 multiple-choice questions. We additionally follow~\cite{buch2022revisiting} to report performance on the ATP-hard subset of the NExT-QA validation set. This subset keeps the hardest QA pairs that can not be solved with one frame, focusing more on long-term temporal reasoning. 

Since each dataset features multiple-choice questions, we utilized accuracy as our evaluation metric.

\subsection{Implementation Details}

We decode all the videos in our experiments at 1 fps and use EVA-CLIP-8B-plus~\cite{sun2024eva} to retrieve the most relevant frames based on the cosine similarity between the generated visual descriptions and the frame features.
For the experiments on EgoSchema, we utilize LaViLa~\cite{zhao2023learning}
as the captioner, which is a clip-based captioning model.
Following~\cite{zhang2023simple}, to ensure zero-shot evaluation, we utilize the LaViLa model retrained on the ego4D data, filtering out the overlapped videos with EgoSchema. 
We sample the video clip for captioning based on the frame index returned by the CLIP retrieval module. 
For NExT-QA, we utilize CogAgent~\cite{hong2023cogagent}
as the captioner. We use GPT-4~\cite{openai2023gpt4} as the LLM for all experiments, the version of GPT is fixed to \textit{gpt-4-1106-preview} to ensure reproducibility.

\subsection{Comparison with State-of-the-arts}

\input{tabs/1_ego}

\emph{VideoAgent} sets new benchmarks, achieving state-of-the-art (SOTA) results on the EgoSchema and NExT-QA datasets, surpassing previous methods significantly while requiring only a minimal number of frames for analysis.

\paragraph{EgoSchema.} As shown in Tables~\ref{tab:ego-public} and~\ref{tab:ego-private}, \emph{VideoAgent} achieves an accuracy of 54.1\% on the EgoSchema full dataset and 60.2\% on a 500-question subset. The full dataset's accuracy was verified by uploading our model's predictions to the official leaderboard, as ground-truth labels are not publicly available. These results not only significantly outperforms the previous SOTA method LLoVi~\cite{zhang2023simple} by 3.8\%, but also achieves comparable performance to advanced proprietary models like Gemini-1.0~\cite{team2023gemini}. Notably, our method requires an average of only 8.4 frames per video --- significantly fewer by 2x to 30x compared to existing approaches.

\paragraph{NExT-QA.} In Table~\ref{tab:nextqa}, we show that \emph{VideoAgent} achieves a 71.3\% accuracy on the NExT-QA full validation set, surpassing the former SOTA, LLoVi~\cite{zhang2023simple}, by 3.6\%. With an average of merely 8.2 frames used per video for zero-shot evaluation, \emph{VideoAgent} consistently outperforms previous supervised and zero-shot methods across all subsets by a large margin, including those testing the model's causal, temporal, and descriptive abilities. Importantly, \emph{VideoAgent} achieves remarkable performance improvements on the more challenging subsets, ATP-hard~\cite{buch2022revisiting}, demonstrating its adeptness at addressing complex long-form video queries.

These results underscore \emph{VideoAgent}'s exceptional effectiveness and efficiency in processing and understanding complex questions from long-form videos.

\input{tabs/2_nextqa}

\subsection{Analysis of Iterative Frame Selection}

\input{tabs/6_abl_round}

One of the key components of \emph{VideoAgent} is its iterative frame selection, which dynamically searches for and aggregates more information until it is sufficient to answer the question, mimicking the human process of understanding videos. We conducted comprehensive analyses and ablation studies to understand this process better.

\paragraph{Frame efficiency.} Our first analysis focused on whether frame selection effectively identifies the informative frames needed to answer a question. This can be measured by frame efficiency: given a fixed number of frames, what model accuracy can be achieved. The hypothesis is that the more informative frames it identifies, the higher the frame efficiency should be. In Figure~\ref{fig:frame_effiency} (left), we plot the accuracy of our method compared to uniform sampling baselines and other previous methods on the EgoSchema 500-question subset. Our method significantly outperforms uniform selection and other baselines at the same number of frames, demonstrating its superiority in frame efficiency. Notably, our method, which uses only 8.4 frames to achieve 60.2\% accuracy, surpasses the baseline that uniformly samples 180 frames to achieve 59.6\% accuracy. This underscores the effectiveness of our method in finding informative frames and reveals that more frames do not always lead to better performance, as irrelevant and noisy information can overwhelm the language model with long contexts and distractions~\cite{liu2024lost,shi2023large}.

\paragraph{Number of rounds.} We also analyzed how the number of iterative rounds affects model performance. In the same Figure~\ref{fig:frame_effiency} (left), we plot the performance across 1-4 rounds and the number of selected frames, achieving accuracies of 53.8\%, 58.6\%, 60.2\%, and 59.8\% with 5, 7.5, 8.4, and 9.9 frames, respectively. The performance improves with additional rounds but saturates at three rounds on the EgoSchema 500-question subset. This indicates that our approach can efficiently find the information needed to answer the question, and beyond a certain point, additional information does not further help in answering the question.

\paragraph{Different question types.} Given that our frame selection process is dynamic, with the language model agent determining whether the information is sufficient, we hypothesized that different question types might require varying amounts of information due to their differing levels of difficulty. We tested this hypothesis on the NExT-QA dataset, which provides annotations for each question type: descriptive tasks, causal reasoning, or temporal reasoning. In Figure~\ref{fig:frame_effiency} (right), we plot the distribution of the number of frames for each type of question. We observed that the average number of frames used ranks as follows: descriptive (5.9 frames), causal (7.1 frames), and temporal (7.8 frames) questions. This aligns with human intuition that descriptive tasks often require fewer frames as initial uniform sampling is usually sufficient, whereas reasoning tasks, especially temporal reasoning, require viewing more frames to accurately answer the question.

\input{tabs/4_abl_component}

\paragraph{Initial Number of Frames.} Before initiating the iterative frame selection process, we uniformly sample $N$ frames to acquaint the language model with the video context. To explore how the number of initially sampled frames influences model performance and the average number of frames utilized, we conduct an ablation study. Specifically, we sample 3, 5, and 8 frames initially on the EgoSchema 500-question subset and report the findings in Table~\ref{tab:init_frame}. The results indicate accuracies of 58.4\%, 60.2\%, and 57.4\% with an average of 6.4, 8.4, and 11.0 frames used, respectively. Starting with 5 frames leads to the highest performance. Furthermore, when comparing our method against uniform sampling with a similar or slightly higher number of frames, we observe accuracies of 54.6\%, 54.8\%, and 55.8\% for 7, 9, and 11 frames, respectively. This comparison again highlights the superior efficiency of our frame selection method.

\paragraph{Self-evaluation.} During the iterative selection process, we perform a self-evaluation to ascertain whether the available information suffices to answer the query. If sufficient, the iteration terminates at this stage. We benchmark this against a baseline method without self-evaluation, where every question is processed through three rounds of iteration. As detailed in Table~\ref{table:segment}, we observe an increase in the average number of frames from 8.4 to 11.8 and a decrease in accuracy from 60.2\% to 59.6\%. These results underscore the efficacy of self-evaluation in determining the adequacy of information, thereby curtailing unnecessary iterations. Notably, gathering more information through additional rounds does not lead to performance improvement but rather results in a marginal decline.

\paragraph{Segment selection.} When it is determined that additional information is required, the input videos are divided into segments. The language model then generates queries specifically tailored to retrieve information within those segments. This approach is contrasted with an alternative strategy that involves generating queries without specifying segments. In Table~\ref{table:segment}, we observe a 3.6\% accuracy degradation when segment selection is disabled. Segment selection improves the model's temporal reasoning capabilities and mitigates the risk of conflating information from disparate segments. This is particularly beneficial for queries such as ``what happens after...?'', where retrieval is only desired from subsequent segments.

\subsection{Ablation of Foundation Models}

Given that \emph{VideoAgent} integrates three foundational model types --- large language model (LLM), visual language model (VLM), and contrastive language-image model (CLIP) --- we conduct a series of ablation studies to evaluate the impact of each component's design on the overall performance of the model.

\input{tabs/5_abl_cap_llm}

\paragraph{LLM.} We initiated our study by evaluating how different LLMs influence the performance of our model, given the pivotal role of LLMs in our methodology, where they function as agents orchestrating the entire process. In Table~\ref{tab:llm}, we compare several state-of-the-art public and proprietary LLMs, including LLaMA-2-70B~\cite{touvron2023llama}, Mixtral-8x7B~\cite{jiang2024mixtral}, GPT-3.5~\cite{ouyang2022training}, and GPT-4~\cite{openai2023gpt4}. Our findings indicate that GPT-4 significantly outperforms its counterparts. However, it is primarily due to its capability in structured prediction. The iterative process employs JSON for output, where accurate JSON parsing is crucial. GPT-4 demonstrates robust performance in generating correct JSON formats, a feat not as consistently achieved by other models, which remains an active research area in LLM~\cite{zheng2023efficiently}.

\paragraph{VLM.} Leveraging GPT-4, a text-only model without visual capabilities, we translate image frames into descriptive captions through VLMs, subsequently feeding these captions to GPT-4. To assess the impact of caption quality produced by various VLMs on performance, we examined three state-of-the-art VLMs: frame-based BLIP-2~\cite{li2023blip2} and CogAgent~\cite{hong2023cogagent}, along with clip-based LaViLa~\cite{zhao2023learning} as presented in Table~\ref{table:captioning}. Our analysis revealed that captions from CogAgent and LaViLa yield similar performances, even though their generated captions have significantly different lengths, while BLIP-2 generated captions are much worse.

\paragraph{CLIP.} CLIP excels in retrieval tasks due to the late-interaction design for image and text features, eliminating the need for recomputing image embeddings for varying queries. We evaluated three versions of CLIP: OpenCLIP ViT-G~\cite{ilharco_gabriel_2021_5143773}, EVA-CLIP-8B~\cite{sun2024eva}, and EVA-CLIP-8B-plus~\cite{sun2024eva}, with the results shown in Table~\ref{tab:clip}. Our findings suggest comparable performance across different CLIP models, indicating that retrieval does not constitute a bottleneck for our methodology.

It's important to note that the main contribution of our research is the introduction of a framework emulating the human process of understanding long-form videos, rather than the employment of any specific model. With the rapid developments of foundational models such as LLMs, VLMs, and CLIPs, our approach can be further improved with the integration of better models, or by adopting a caption-free methodology by replacing GPT-4 with GPT-4V. We hope our work sheds light on future work in this direction.

\subsection{Case Studies}

\begin{figure}[!tb]
    \centering
    \includegraphics[width=\linewidth]{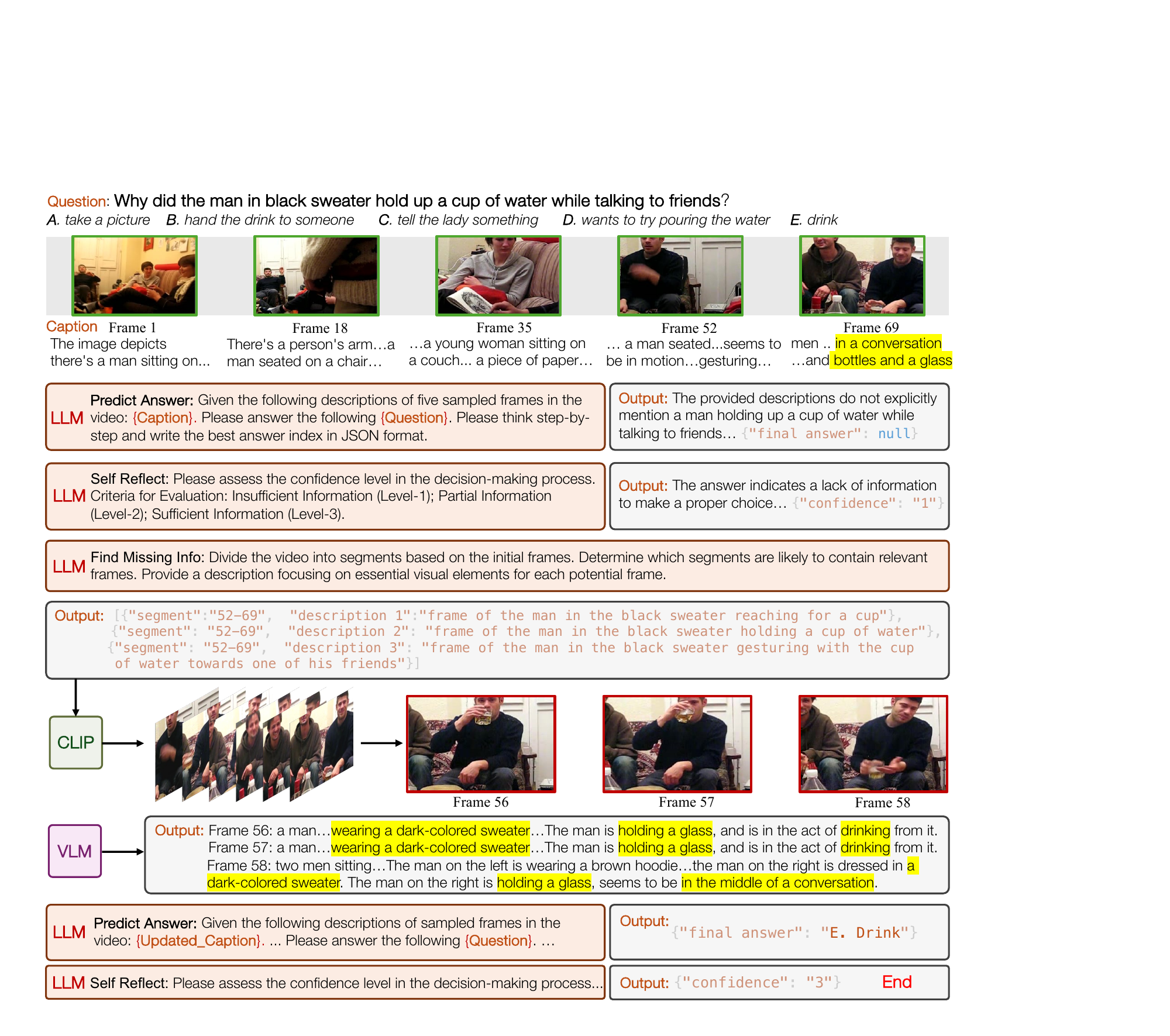}
    \vspace{-1.5em}
    \caption{\emph{Case study on NExT-QA.} \emph{VideoAgent} accurately identifies missing information in the first round, bridges the information gap in the second round, and thereby makes the correct prediction. }
    \label{fig:case1}
    \vspace{-1em}
\end{figure}

\begin{figure}[!tb]
    \centering
    \includegraphics[width=\linewidth]{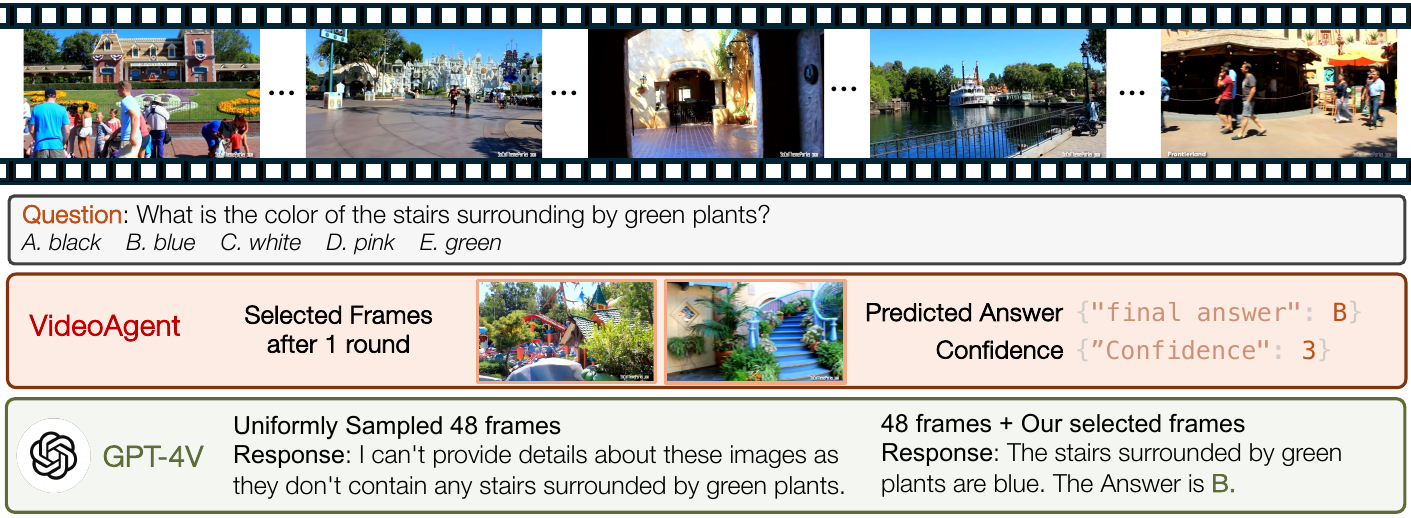}
    \vspace{-1.5em}
    \caption{\emph{Case study on hour-long videos.} \emph{VideoAgent} accurately identifies the key frame during the second iteration, subsequently making an accurate prediction. Conversely, GPT-4V, when relying on 48 uniformly sampled frames up to its maximum context length, does not get successful prediction. However, by integrating the frame pinpointed by \emph{VideoAgent}, GPT-4V is able to correctly answer the question. }
    \vspace{-1em}
    \label{fig:case2}
\end{figure}

We present several case studies to demonstrate the capability of \emph{VideoAgent} in understanding long-form videos.

\paragraph{Questions from NExT-QA~\cite{xiao2021next}.} In Figure~\ref{fig:case1}, we illustrate an instance from NExT-QA solved in two iterations. The question is asking why the man holds up a cup of water when talking to friends. \emph{VideoAgent} accurately identify missing information (although the cup is visible in frame 69, it does not reveal the man is holding it). It then determines what additional information is required (frame of the man in the black sweater holding a cup of water). Finally, it utilizes CLIP to retrieve this detail (the man is holding a glass and is in the act of drinking from it) and feel confident about its answer. 

\paragraph{Hour-long videos.} Given that both NExT-QA and EgoSchema videos span only a few minutes, Figure~\ref{fig:case2} shows how \emph{VideoAgent} can accurately solve hour-long videos from YouTube\footnote{\url{https://www.youtube.com/watch?v=H9Y5_X1sEEA}}. The question is about figuring out the color of the stairs surrounding by green plants, which only occupy a small portion of the video. 
\emph{VideoAgent} efficiently identifies the necessary information and answers questions within only two iterations and seven frames, outperforming state-of-the-art models like GPT-4V. Notably, GPT-4V struggles with uniform sampling across its maximum context length of 48 images. However, when GPT-4V is provided with the frame pinpointed by \emph{VideoAgent}, it can successfully answer the question. This underscores the potential of enhancing GPT-4V's capabilities in video understanding by integrating our approach.

In conclusion, \emph{VideoAgent} is ready to tackle real-world video understanding challenges, surpassing traditional methods reliant on one-round sparse or dense sampling.

%% file: tabs/1_ego.tex
\begin{figure}[tb]
	\begin{minipage}{\textwidth}
	  \begin{minipage}[t]{0.495\textwidth}
		\begin{threeparttable}
	\resizebox{1\textwidth}{!}{\setlength\tabcolsep{2pt}
    \renewcommand\arraystretch{1.03}
    \begin{tabular}{rl|ccc}
    \toprule
    \multicolumn{2}{c|}{Method} & Frames & Subset & \textbf{Full} \\ 
    \midrule
FrozenBiLM~\cite{yang2022frozenbilm}\!&\pub{NeurIPS2022} & 90 & - & 26.9 \\
InternVideo~\cite{wang2022internvideo}\!&\pub{arXiv2022.12} &  90 & - &  32.1 \\
ImageViT~\cite{papalampidi2023simple}\!&\pub{arXiv2023.12} & 16 & 40.8 & 30.9 \\
ShortViViT$_{loc}$~\cite{papalampidi2023simple}\!&\pub{arXiv2023.12} & 32 & 49.6 & 31.3 \\  
LongViViT~\cite{papalampidi2023simple}\!&\pub{arXiv2023.12} & 256 & 56.8 & 33.3 \\  
SeViLA~\cite{yu2023self}\!&\pub{NeurIPS2023} & 32 & 25.7 & 22.7 \\ 
Vamos~\cite{wang2023vamos}\!&\pub{arXiv2023.11} & - & - & 48.3 \\
LLoVi~\cite{zhang2023simple}\!&\pub{arXiv2024.2} & 180 & 57.6 & 50.3 \\
MC-ViT-L~\cite{balavzevic2024memory}\!&\pub{arXiv2024.2} & 128+ & 62.6 & 44.4 \\
\midrule 
VideoAgent&\footnotesize{(\textbf{\texttt{ours}})} & \textbf{8.4} & 60.2 & \textbf{54.1} \\ 
    \bottomrule
    \vspace{-1.85em}
    \end{tabular}
		}
	\end{threeparttable}
 \makeatletter\def\@captype{table}\makeatother\caption{{\emph{Results on EgoSchema compared to public models.} Full-set results are obtained from the official leaderboard.\label{tab:ego-public}}}
	  \end{minipage}
   	  \begin{minipage}[t]{0.005\textwidth}
	  ~~~~~~
	  \end{minipage}
	  \begin{minipage}[t]{0.495\textwidth}
\begin{threeparttable}
\resizebox{1\textwidth}{!}{\setlength\tabcolsep{4pt}
	\renewcommand\arraystretch{1.03}
    \begin{tabular}{rl|cc}
    \toprule
    \multicolumn{2}{c|}{Model} & Subset & \textbf{Full}\\ 
    \midrule
    Random Chance && 20.0 & 20.0 \\
    Bard only (blind)~\cite{balavzevic2024memory}\!&\pub{2023.3} & 27.0 & 33.2 \\
    Bard + ImageViT~\cite{papalampidi2023simple}\!&\pub{2023.3} & 35.0 & 35.0\\
    Bard + ShortViViT~\cite{papalampidi2023simple}\!&\pub{2023.3} & 42.0 & 36.2\\
    Bard + PALI~\cite{papalampidi2023simple}\!&\pub{2023.3} & 44.8 & 39.2 \\
    GPT-4 Turbo (blind)~\cite{balavzevic2024memory}\!&\pub{2023.4} & 31.0 & 30.8 \\
    GPT-4V~\cite{balavzevic2024memory}\!&\pub{2023.9} & 63.5 & 55.6 \\
    Gemini 1.0 Pro~\cite{team2023gemini}\!&\pub{2023.12} & - & 55.7 \\
    \midrule
    VideoAgent&\footnotesize{(\textbf{\texttt{ours}})} & 60.2 & 54.1 \\ 
    \bottomrule
    \end{tabular}
    }
\end{threeparttable}
\makeatletter\def\@captype{table}\makeatother\caption{\emph{Results on EgoSchema compared to large-scale proprietary models.}}\label{tab:ego-private}
	  \end{minipage}
		\end{minipage}
 \vspace{-1em}
        \label{tab:ego}
\end{figure}

%% file: tabs/2_nextqa.tex
\begin{table}[t]
    \centering
    \resizebox{1\textwidth}{!}{
    \setlength\tabcolsep{4pt}
    \renewcommand\arraystretch{0.8}
    \begin{tabular}{rl|cccc|ccc}
    \toprule
    & & \multicolumn{4}{c|}{Val} & \multicolumn{3}{c}{ATP-hard subset}  \\
    \multicolumn{2}{c|}{\multirow{-2}{*}{Methods}}  & Acc@C & Acc@T & Acc@D &  Acc@All & Acc@C & Acc@T &  Acc@All  \\ 
    \midrule
    \multicolumn{9}{c}{\textit{Supervised}} \\
    VFC~\cite{yang2021just}\!&\pub{ICCV2021}  & 49.6 & 51.5 & 63.2 & 52.3 & - & - & - \\
    ATP~\cite{buch2022revisiting}\!&\pub{CVPR2022}  & 53.1 & 50.2 & 66.8 & 54.3 & 38.4 & 36.5 & 38.8 \\
    MIST~\cite{gao2023mist}\!&\pub{CVPR2023}  & 54.6 &  56.6 & 66.9 &  57.2 & - & - & -\\
    GF~\cite{bai2024glance}\!&\pub{NeurIPS2023}  & 56.9 & 57.1 &  70.5 & 58.8 &48.7 & 50.3 & 49.3 \\
    CoVGT~\cite{xiaovgt}\!&\pub{TPAMI2023}   &59.7 & 58.0& 69.9 & 60.7 & - &-&- \\
    SeViT~\cite{kim2023semi}\!&\pub{arXiv2023.1} & 54.0 & 54.1 & 71.3 & 56.7 & 43.3 & 46.5 & - \\
    HiTeA~\cite{ye2023hitea}\!&\pub{ICCV2023} & 62.4 & 58.3 & 75.6 & 63.1 & 47.8 & 48.6 &  - \\
    \midrule
    \multicolumn{9}{c}{\textit{Zero-shot}} \\
    VFC~\cite{momeni2023verbs}\!&\pub{ICCV2023} & 51.6 & 45.4 & 64.1 & 51.5 & 32.2 & 30.0 & 31.4 \\
    InternVideo~\cite{wang2022internvideo}\!&\pub{arXiv2022.12} & 43.4 & 48.0 & 65.1 & 49.1 & - & - & - \\
    AssistGPT~\cite{gao2023assistgpt}\!&\pub{arXiv2023.6} & 60.0 & 51.4 & 67.3 & 58.4 & - & - & - \\
    ViperGPT~\cite{suris2023vipergpt}\!&\pub{ICCV2023} & - & - & - & 60.0 & - & - & - \\
    SeViLA~\cite{yu2023self}\!&\pub{NeurIPS2023} & 61.3 & 61.5 & 75.6 & 63.6 & - & - & -  \\
    LLoVi~\cite{zhang2023simple}\!&\pub{arXiv2024.2} & 69.5 & 61.0 & 75.6 & 67.7 & - & - & - \\
    \midrule
    VideoAgent&\footnotesize{(\textbf{\texttt{ours}})} & \textbf{72.7} & \textbf{64.5} & \textbf{81.1} & \textbf{71.3} & \textbf{57.8} & \textbf{58.8} & \textbf{58.4}  \\
    \bottomrule
    \end{tabular}
    }
    \caption{\emph{Results on NExT-QA compared to the state of the art.} C, T, and D are causal, temporal, and descriptive subsets, respectively.}
    \label{tab:nextqa}
    \vspace{-2em}
\end{table}

%% file: tabs/6_abl_round.tex
\begin{figure}[tb]
	\begin{center}
		\includegraphics[width=0.49\linewidth]{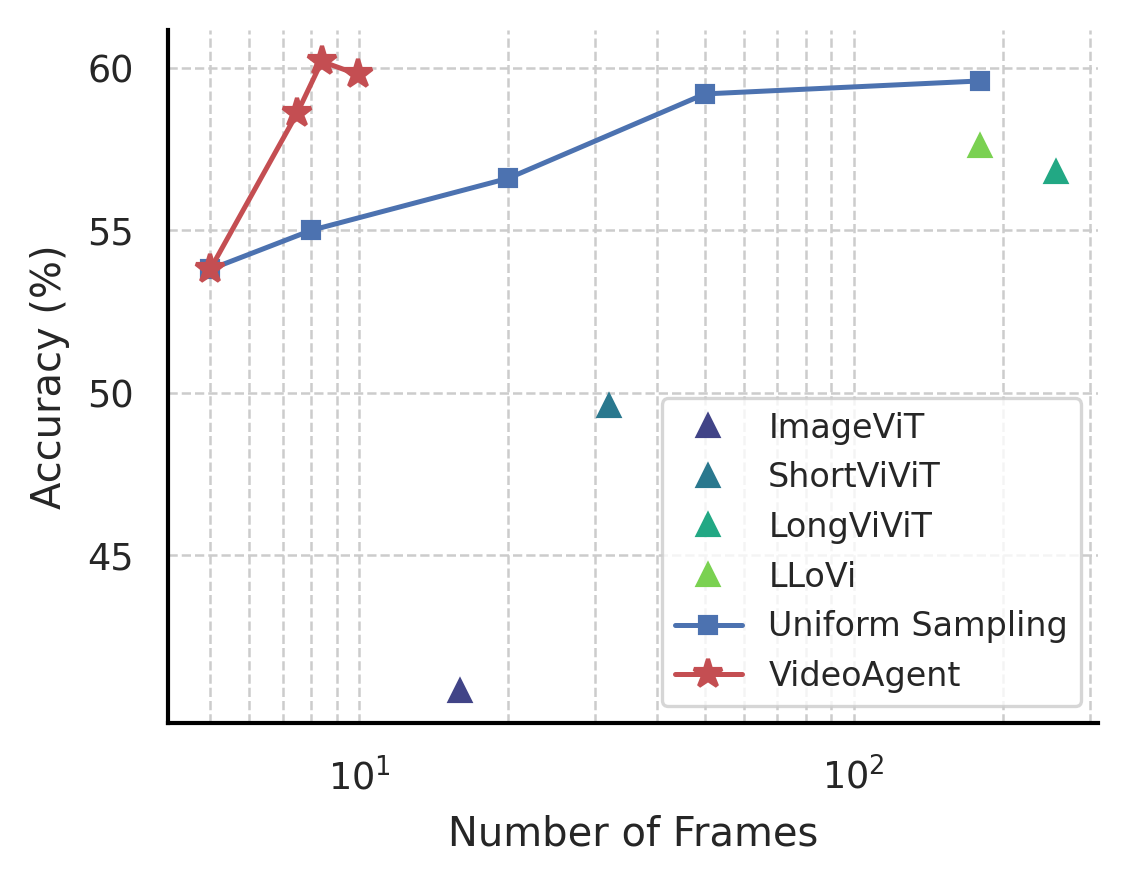}
        \includegraphics[width=0.49\linewidth]{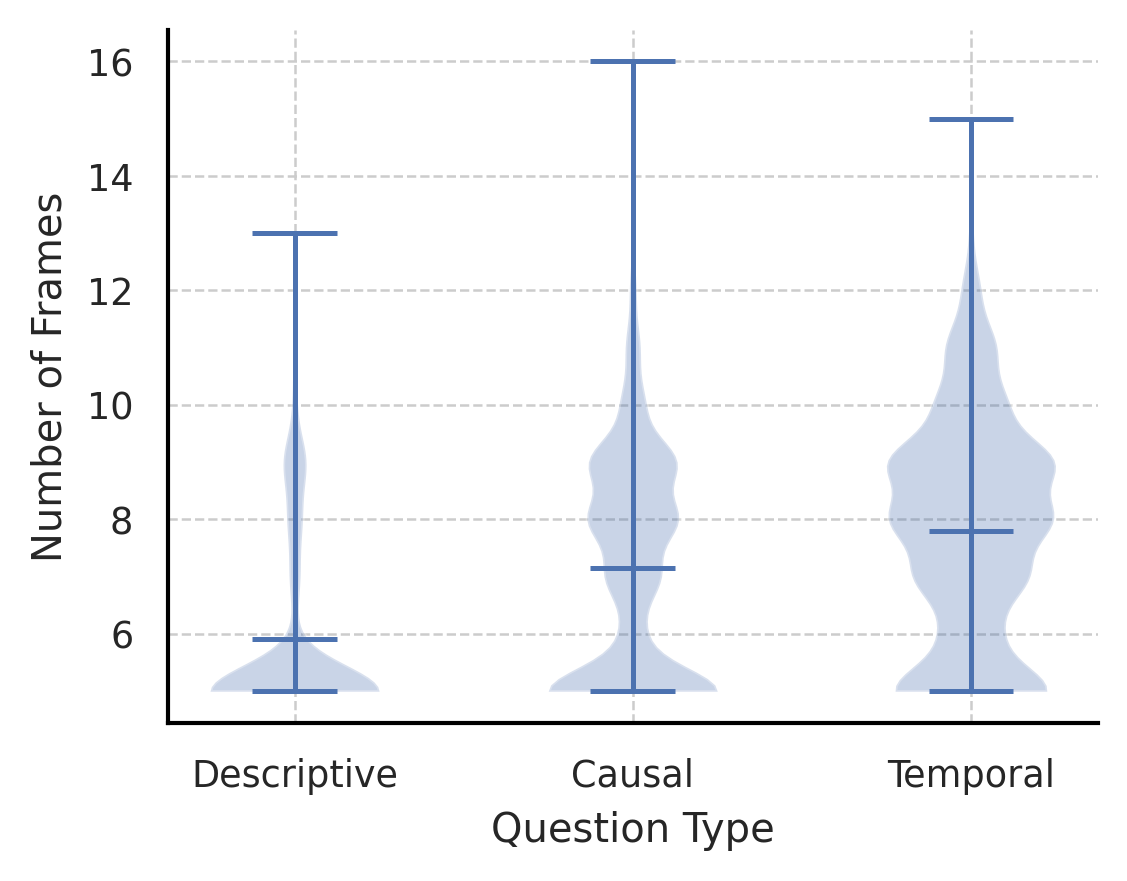}
	\end{center}
 \vspace{-2em}
	\caption{\emph{(Left) Frame efficiency compared to uniform sampling and previous methods.} X-axis is in log scale. Our method achieves exceptional frame efficiency for long-form video understanding. \emph{(Right) Number of frames for different types of NExT-QA questions.} Min, mean, max, distribution are plotted. \emph{VideoAgent} selects more frames on questions related to temporal reasoning than causal reasoning and descriptive questions.}
 \vspace{-1em}
	\label{fig:frame_effiency}
\end{figure}

%% file: tabs/4_abl_component.tex
\begin{figure*}[tb]
	\begin{minipage}{\textwidth}
 
	  \begin{minipage}[t]{0.47\textwidth}
		\begin{threeparttable}
	\resizebox{1\textwidth}{!}{\setlength\tabcolsep{5pt}
    \renewcommand\arraystretch{1.01}
\begin{tabular}{l|ccc}
\toprule
 \multirow{2}{*}{Uniform}  & Uni-7 & Uni-9 & Uni-11  \\
 & 54.6 & 54.8 & 55.8  \\
\midrule
\multirow{2}{*}{Ours} & 3$\rightarrow$6.4 &5$\rightarrow$8.4 & 8$\rightarrow$11.0  \\ 
& \textbf{58.4} & \textbf{60.2} & \textbf{57.4}  \\
\bottomrule
\end{tabular}
}
\end{threeparttable}
\vspace{-1em}
\makeatletter\def\@captype{table}\makeatother\caption{\emph{Ablation of initial number of uniformly sampled frames.\label{tab:init_frame}}}
	  \end{minipage}
   	  \begin{minipage}[t]{0.005\textwidth}
	  ~~~~~~
	  \end{minipage}
	  \begin{minipage}[t]{0.52\textwidth}
\begin{threeparttable}
\resizebox{1\textwidth}{!}{\setlength\tabcolsep{4pt}
	\renewcommand\arraystretch{1.03}
\begin{tabular}{ccc}
\toprule
\ Method & Frames & Acc  \\ 
\midrule
Ours w/o Seg. Selection & 7.5 & 56.6 \\
Ours w/o Self-Evaluation & 11.8 & 59.6 \\
Ours & 8.4 & \textbf{60.2}  \\
\bottomrule
\end{tabular}
    }
\end{threeparttable}
  \vspace{-1em}
\makeatletter\def\@captype{table}\makeatother\caption{\emph{Ablation of segment selection and self-evaluation.}}\label{table:segment}
	  \end{minipage}
		\end{minipage}
        \label{tab:component}
        \vspace{-1em}
	\end{figure*}

%% file: tabs/5_abl_cap_llm.tex
\begin{figure*}[tb]
	\begin{minipage}{\textwidth}
 	  \begin{minipage}[t]{0.26\textwidth}
\begin{threeparttable}
\resizebox{1\textwidth}{!}{
			\setlength\tabcolsep{5.3pt}
				\renewcommand\arraystretch{1.045}
        \begin{tabular}{lccccc}
    \toprule
    LLM & Model Size & Acc. (\%)\\ 
    \midrule
    Mistral-8x7B & 70B & 37.8 \\
    Llama2-70B & 70B & 45.4 \\
    GPT-3.5 & N/A & 48.8 \\
    GPT-4 & N/A & \textbf{60.2} \\
    \bottomrule
    \end{tabular}
    		}
\end{threeparttable}
  \vspace{-1em}
\makeatletter\def\@captype{table}\makeatother\caption{\scriptsize{\emph{LLM ablation.}}}\label{tab:llm}
	  \end{minipage}
	  \begin{minipage}[t]{0.35\textwidth}
		\begin{threeparttable}
			\resizebox{1\textwidth}{!}{
			\setlength\tabcolsep{5.3pt}
				\renewcommand\arraystretch{1.03}
    \begin{tabular}{lccccc}
    \toprule
    Captioner & Type & \# Words & Acc. (\%)\\ 
    \midrule
    BLIP-2 & Frame-based & 8.5 & 52.4 \\
    LaViLa & Clip-based & 7.2 & 60.2 \\
    CogAgent & Frame-based & 74.2 & \textbf{60.8}  \\
    \bottomrule
    \end{tabular}
		}
	\end{threeparttable}
 \vspace{-1em}
 \makeatletter\def\@captype{table}\makeatother\caption{{\scriptsize{\emph{VLM ablation.}}\label{table:captioning}}}
	  \end{minipage}
	  \begin{minipage}[t]{0.35\textwidth}
\begin{threeparttable}
\resizebox{1\textwidth}{!}{
			\setlength\tabcolsep{5.3pt}
				\renewcommand\arraystretch{1.045}
        \begin{tabular}{lccccc}
    \toprule
    CLIP & Model Size & Resolution & Acc. (\%)\\ 
    \midrule
    OpenCLIP ViT-G & 1B & 224 & 59.2 \\
    EVA-CLIP-8B & 8B & 224 & 59.4 \\
    EVA-CLIP-8B-plus & 8B & 448 & \textbf{60.2} \\
    \bottomrule
    \end{tabular}
    		}
\end{threeparttable}
  \vspace{-1em}
\makeatletter\def\@captype{table}\makeatother\caption{\scriptsize{\emph{CLIP ablation.}}}\label{tab:clip}
	  \end{minipage}
		\end{minipage}
 \vspace{-1em}
	\end{figure*}

%% file: secs/5_conclusion.tex
\section{Conclusion}

In this work, we introduce \emph{VideoAgent}, a system that employs a large language model as an agent to mirror the human cognitive process for understanding long-form videos. \emph{VideoAgent} effectively searches and aggregates information through a multi-round iterative process. It demonstrates exceptional effectiveness and efficiency in long-form video understanding, as evidenced by both quantitative and qualitative studies on various datasets.

%% file: secs/X_suppl.tex
\clearpage
\setcounter{page}{1}

\title{\emph{VideoAgent}: Long-form Video Understanding with Large Language Model as Agent} 
\titlerunning{\emph{VideoAgent}}
\author{Xiaohan Wang$^\star$ \and Yuhui Zhang$^\star$ \and Orr Zohar \and Serena Yeung-Levy}
\authorrunning{Wang et al.}
\institute{Stanford University \\
\email{\{xhanwang,yuhuiz,orrzohar,syyeung\}@stanford.edu}}
\renewcommand{\thefootnote}{\ensuremath{\dagger}}
{
  \renewcommand{\thefootnote}
    {\fnsymbol{footnote}}
  \footnotetext[1]{Equal contribution. Project page: \scriptsize{\url{https://wxh1996.github.io/VideoAgent-Website/}}.}
}

\maketitle

This document provides more details of our approach and additional experimental results, organized as follows:
\begin{itemize}
	\vspace{-3pt}
	\setlength{\itemsep}{0pt}
	\setlength{\parsep}{0pt}
	\setlength{\parskip}{0pt}
 	\item \S~\ref{sec:s3} Run-time Analysis of CLIP
	\item \S~\ref{sec:s4} Additional Implementation Details
	\item \S~\ref{sec:s5} Prompts for GPT-4
\end{itemize}

\section{Run-time Analysis of CLIP}\label{sec:s3}

While CLIP in \emph{VideoAgent} may see up to all the frames within the video and VLM only captions a few frames, using CLIP is computationally efficient and negligible compared to using an LLM or VLM for several reasons. Firstly, CLIP's feature computation involves just a single feed-forward process. Secondly, CLIP employs an image-text late interaction architecture, enabling the caching and reusing of image frame features across different text queries. 

Consider the scenario where the computation of CLIP features requires $x$ seconds per image and text, while VLM captioning necessitates $y$ seconds per image, and LLM computation consumes $z$ seconds per round. For a video comprising $N$ frames, suppose the \emph{VideoAgent} selectively processes $n$ frames out of $N$ across $t$ rounds of iterations. In this context, the time to compute CLIP features for images and texts amounts to $Nx$ and $nx$ seconds respectively. The generation of VLM captions requires $ny$ seconds, and LLM operations total $tz$ seconds. Consequently, the proportion of time dedicated to computing CLIP features relative to the overall computation time is approximated by $\frac{Nx+nx}{Nx+nx+ny+tz}$.

In practice, utilizing OpenCLIP ViT-G as CLIP, CogAgent as VLM, GPT-4 as LLM, with an A6000 GPU and the EgoSchema dataset, we find that $N=180$, $n=8.4$, $x=0.02$, $y=20$, $z=10$, and $t=3$. Therefore, the formula simplifies to $\frac{180 \times 0.02 + 8.4 \times 0.02}{180 \times 0.02 + 8.4 \times 0.02 + 8.4 \times 20 + 3 \times 10}$, which evaluates to 1.9\%. This demonstrates that, under these conditions, the computation of CLIP features represents a small fraction, specifically 1.9\%, of the total computational effort.

It should also be noted that the estimation above represents an upper bound. In practice, our segment-level retrieval approach necessitates computing features only within specified segments, rather than across all $N$ frames, which further enhances efficiency.

\section{Additional Implementation Details}\label{sec:s4}

\paragraph{Details of CogAgent.} For the experiments on NExT-QA~\cite{xiao2021next}, we utilize CogAgent~\cite{zhao2023learning}
as the captioner, which is a frame-based captioning model. CogAgent has 18B parameters with input image resolution 1120 $\times$ 1120.

\paragraph{Details of LaViLa.} For the experiments on EgoSchema~\cite{mangalam2023egoschema}, we utilize LaViLa~\cite{zhao2023learning} as the captioner, which is a clip-based captioning model. LaViLA takes input clip with resolution at 4 $\times$ 336  $\times$ 336. Following~\cite{zhang2023simple}, to ensure zero-shot evaluation, we utilize the LaViLa model retrained on the ego4D data, filtering out the overlapped videos with EgoSchema. 

\paragraph{Details of CLIP.} We employ the EVA-CLIP-8B-plus~\cite{sun2024eva} model for frame retrieval, a state-of-the-art CLIP model that includes a vision encoder with 7.5 billion parameters and a text encoder with 0.7 billion parameters. This model processes images at a resolution of 448 $\times$ 448 and produces output features with a dimensionality of 1280.

\section{Prompts for GPT-4}\label{sec:s5}

The specific prompts utilized for GPT-4 to predict answer, self reflect, and find missing information are shown in Figures~\ref{fig:prompt1},~\ref{fig:prompt2}, and~\ref{fig:prompt3}, respectively.

\begin{figure*}[htb]
\begin{minted}[breaklines,fontsize=\tiny,escapeinside=||]{text}
Given a video that has 180 frames, the frames are decoded at 1 fps. Given the following descriptions of the sampled frames in the video:
|\textcolor{orange}{\{'frame 1': '#C C rolls the dough on the table with both hands.', 'frame 28': '#C C puts the dough in the dough roller', 'frame 45': '#C C walks to the doughs on the work table.', 'frame 76': '#C C picks dough from the baking tray', 'frame 90': '#C C picks up dough', 'frame 111': '#C C picks dough from the tray of doughs with both hands. ', 'frame 135': '#C C throws the dough on the dough roller', 'frame 171': '#C C places the dough on the baking tray', 'frame 180': '#C C rolls the dough on the baking table with his hands.'\}}|
Please answer the following question:
```
|\textcolor{orange}{How would you briefly describe the sequential order of the process that c performed on the dough, from initial handling to final placement on the tray?}|
|\textcolor{orange}{0. Initially, c first carefully rolled the dough on the flour, then smoothly rolled it on the table, and finally, gently placed it in the awaiting tray.}|
|\textcolor{orange}{1. C first placed the dough in the tray, then rolled it on the table, and finally rolled it on the flour.}|
|\textcolor{orange}{2. Initially, c carefully rolled the dough on the table, then skillfully placed it in the tray, and ultimately, gently rolled it on the flour.}|
|\textcolor{orange}{3. Initially, c first carefully placed the dough in the tray, then gently rolled it on the flour, and ultimately smoothly rolled it on the table.}|
|\textcolor{orange}{4. C first rolled the dough on the table, then rolled it on the flour, and finally placed it in the tray.}|
```
Please think step-by-step and write the best answer index in Json format {'final_answer': 'xxx'}. Note that only one answer is returned for the question.
\end{minted}
    \caption{\textbf{The prompt to predict answer.} We instantiate the prompt with a real video and question (highlighted in orange).}
    \label{fig:prompt1}
\end{figure*}

\begin{figure*}[htb]
\begin{minted}[breaklines,fontsize=\tiny,escapeinside=||]{text}
Please assess the confidence level in the decision-making process.
The provided information is as as follows,
Given a video that has 180 frames, the frames are decoded at 1 fps. Given the following descriptions of the sampled frames in the video:
|\textcolor{orange}{\{'frame 1': '#C C rolls the dough on the table with both hands.', 'frame 28': '#C C puts the dough in the dough roller', 'frame 45': '#C C walks to the doughs on the work table.', 'frame 76': '#C C picks dough from the baking tray', 'frame 90': '#C C picks up dough', 'frame 111': '#C C picks dough from the tray of doughs with both hands. ', 'frame 135': '#C C throws the dough on the dough roller', 'frame 171': '#C C places the dough on the baking tray', 'frame 180': '#C C rolls the dough on the baking table with his hands.'\}}|
Please answer the following question:
```
|\textcolor{orange}{How would you briefly describe the sequential order of the process that c performed on the dough, from initial handling to final placement on the tray?}|
|\textcolor{orange}{0. Initially, c first carefully rolled the dough on the flour, then smoothly rolled it on the table, and finally, gently placed it in the awaiting tray.}|
|\textcolor{orange}{1. C first placed the dough in the tray, then rolled it on the table, and finally rolled it on the flour.}|
|\textcolor{orange}{2. Initially, c carefully rolled the dough on the table, then skillfully placed it in the tray, and ultimately, gently rolled it on the flour.}|
|\textcolor{orange}{3. Initially, c first carefully placed the dough in the tray, then gently rolled it on the flour, and ultimately smoothly rolled it on the table.}|
|\textcolor{orange}{4. C first rolled the dough on the table, then rolled it on the flour, and finally placed it in the tray.}|
```
Please think step-by-step and write the best answer index in Json format {'final_answer': 'xxx'}. Note that only one answer is returned for the question.
The decision making process is as follows,
|\textcolor{orange}{Based on the sequence of sampled frames provided, the following sequential order of the process can be described:}|

|\textcolor{orange}{- C rolls the dough on the table with both hands. (frame 1 & frame 180)}|
|\textcolor{orange}{- C puts the dough in the dough roller. (frame 28 & frame 135)}|
|\textcolor{orange}{- C walks to the doughs on the work table. (frame 45, context not clear enough to make this a major step)}|
|\textcolor{orange}{- C picks dough from the baking tray. (frame 76)}|
|\textcolor{orange}{- C picks up dough. (frame 90, likely a continuation of picking it from the tray based on the next frame)}|
|\textcolor{orange}{- C picks dough from the tray of doughs with both hands. (frame 111, this confirms the action started at frame 76)}|
|\textcolor{orange}{- C places the dough on the baking tray. (frame 171)}|

|\textcolor{orange}{From the above sequence, the most accurate description is that C first rolls the dough on the table, then puts it through the dough roller, and finally places it on the baking tray. Thus, the correct sequential order, represented by the best answer, should be:}|

|\textcolor{orange}{4. C first rolled the dough on the table, then rolled it on the flour, and finally placed it in the tray.}|

|\textcolor{orange}{Now to provide the answer in the requested JSON format:}|

|\textcolor{orange}{```}|
|\textcolor{orange}{\{'final\_answer': '4'\}}|
|\textcolor{orange}{```}|
Criteria for Evaluation:
Insufficient Information (Confidence Level: 1): If information is too lacking for a reasonable conclusion.
Partial Information (Confidence Level: 2): If information partially supports an informed guess.
Sufficient Information (Confidence Level: 3): If information fully supports a well-informed decision.
Assessment Focus:
Evaluate based on the relevance, completeness, and clarity of the provided information in relation to the decision-making context.
Please generate the confidence with JSON format {'confidence': 'xxx’}
\end{minted}
    \caption{\textbf{The prompt to self reflect.} We instantiate the prompt with a real video and question (highlighted in orange).}
    \label{fig:prompt2}
\end{figure*}

\begin{figure*}[htb]
\begin{minted}[breaklines,fontsize=\tiny,escapeinside=||]{text}
Given a video that has 180 frames, the frames are decoded at 1 fps. Given the following descriptions of five uniformly sampled frames in the video:
|\textcolor{orange}{\{'frame 1': '#C C rolls the dough on the table with both hands.', 'frame 45': '#C C walks to the doughs on the work table.', 'frame 90': '#C C picks up dough', 'frame 135': '#C C throws the dough on the dough roller', 'frame 180': '#C C rolls the dough on the baking table with his hands.'\}}|
To answer the following question:
```
|\textcolor{orange}{How would you briefly describe the sequential order of the process that c performed on the dough, from initial handling to final placement on the tray?}|
|\textcolor{orange}{0. Initially, c first carefully rolled the dough on the flour, then smoothly rolled it on the table, and finally, gently placed it in the awaiting tray.}|
|\textcolor{orange}{1. C first placed the dough in the tray, then rolled it on the table, and finally rolled it on the flour.}|
|\textcolor{orange}{2. Initially, c carefully rolled the dough on the table, then skillfully placed it in the tray, and ultimately, gently rolled it on the flour.}|
|\textcolor{orange}{3. Initially, c first carefully placed the dough in the tray, then gently rolled it on the flour, and ultimately smoothly rolled it on the table.}|
|\textcolor{orange}{4. C first rolled the dough on the table, then rolled it on the flour, and finally placed it in the tray.}|
```
However, the information in the initial five frames is not suffient.
Objective:
Our goal is to identify additional frames that contain crucial information necessary for answering the question. These frames should not only address the query directly but should also complement the insights gleaned from the descriptions of the initial five frames.
To achieve this, we will:
1. Divide the video into four segments based on the intervals between the initial five frames.
2. Determine which segments are likely to contain frames that are most relevant to the question. These frames should capture key visual elements, such as objects, humans, interactions, actions, and scenes, that are supportive to answer the question.
For each frame identified as potentially relevant, provide a concise description focusing on essential visual elements. Use a single sentence per frame. If the specifics of a segment's visual content are uncertain based on the current information, use placeholders for specific actions or objects, but ensure the description still conveys the segment's relevance to the query.
Select multiple frames from one segment if necessary to gather comprehensive insights.
```
{'frame_descriptions': [{'segment_id': '1/2/3/4', 'duration': 'xxx - xxx', 'description': 'frame of xxx'}, {'segment_id': '1/2/3/4', 'duration': 'xxx - xxx', 'description': 'frame of xxx'}, {'segment_id': '1/2/3/4', 'duration': 'xxx - xxx', 'description': 'frame of xxx'}]}
\end{minted}
    \caption{\textbf{The prompt to find missing information.} We instantiate the prompt with a real video and question (highlighted in orange).}
    \label{fig:prompt3}
\end{figure*}